\documentclass[sigconf, screen]{acmart}

\usepackage{multirow}
\usepackage{balance}
\usepackage{graphicx}
\usepackage{subcaption}
\usepackage{booktabs}
\usepackage{multirow}
\usepackage[most]{tcolorbox}
\usepackage{makecell}
\usepackage{array}
\usepackage{colortbl}
\usepackage{xcolor}
\usepackage{caption}
\usepackage{rotating}
\usepackage{pifont}
\usepackage{tikz}
\usepackage{enumitem} 

\newcommand{\yunbo}[1]{\textcolor{black}{#1}}
\newcommand{\tool}{CLIP-Enhance}

\definecolor{lightblue}{HTML}{B7E0FF}

\AtBeginDocument{%
  }

\copyrightyear{2025}
\acmYear{2025}
\setcopyright{cc}
\setcctype{by}
\acmConference[MM '25]{Proceedings of the 33rd ACM International Conference
on Multimedia}{October 27--31, 2025}{Dublin, Ireland}
\acmBooktitle{Proceedings of the 33rd ACM International Conference on
Multimedia (MM '25), October 27--31, 2025, Dublin,
Ireland}\acmDOI{10.1145/3746027.3755748}
\acmISBN{979-8-4007-2035-2/2025/10}

\begin{document}

\title{Do Existing Testing Tools Really Uncover Gender Bias in Text-to-Image Models?}

\author{Yunbo Lyu}
\affiliation{%
  \institution{Singapore Management University}
  \country{Singapore}}
\email{yunbolyu@smu.edu.sg}

\author{Zhou Yang$^*$}
\affiliation{%
  \institution{University of Alberta}
  \authornote{Corresponding author.}
  \city{Edmonton}
  \country{Canada}}
\email{zy25@ualberta.ca}

\author{Yuqing Niu}
\affiliation{%
  \institution{Singapore Management University}
  \country{Singapore}}
\email{yuqingniu@smu.edu.sg}

\author{Jing Jiang}
\affiliation{%
  \institution{Australian National University}
  \city{Canberra}
  \country{Australia}}
\email{jing.jiang@anu.edu.au}

\author{David Lo}
\affiliation{%
  \institution{Singapore Management University}
  \country{Singapore}}
\email{davidlo@smu.edu.sg}

\renewcommand{\shortauthors}{Lyu et al.}

\begin{abstract}

  Text-to-Image (T2I) models have recently gained significant attention due to their ability to generate high-quality images and are consequently used in a wide range of applications. 
  However, there are concerns about the gender bias of these models.
  Previous studies have shown that T2I models \yunbo{can perpetuate or even amplify gender stereotypes when provided with neutral text prompts (e.g., `a photo of a CEO' is often associates with male images, while `a photo of nurse' is often associates with female images).}
  Researchers have proposed automated gender bias uncovering detectors for T2I models, but a crucial gap exists: \textit{no existing work comprehensively compares the various detectors and understands how the gender bias detected by them deviates from the actual situation.}

  This study addresses this gap by \yunbo{validating} previous gender bias detectors using a \yunbo{manually labeled} dataset and comparing how the bias identified by various detectors deviates from the actual bias in T2I models, as verified by manual confirmation.
  We create a dataset consisting of 6,000 images generated from three cutting-edge T2I models, Stable Diffusion XL, Stable Diffusion 3, and Dreamlike Photoreal 2.0.
  During the human-labeling process, we find that all three T2I models generate a portion (12.48\% on average) of low-quality images (e.g., generate images with no face present), where human annotators cannot determine the gender of the person.

  Our analysis reveals that all three T2I models show a preference for generating male images, with SDXL being the most biased.
  Additionally, images generated using prompts containing professional descriptions (e.g., lawyer or doctor) show the most bias.
  We evaluate seven gender bias detectors and find that none fully capture the actual level of bias in T2I models, with some detectors overestimating bias by up to 26.95\%.
  We further investigate the causes of inaccurate estimations, highlighting the limitations of detectors in dealing with low-quality images.
  Based on our findings, we propose an enhanced detector called \tool{}, which most accurately measures the gender bias in T2I models, with a difference of only 0.47\%-1.23\%, and most effectively filters out 82.91\% of low-quality images.
  \footnote{This paper potentially contains offensive information for some groups.}
  We have made our dataset and code publicly available.
  \footnote{https://doi.org/10.6084/m9.figshare.27377649.v1}
\end{abstract}



    




\received{11 April 2025}
\received[accepted]{4 July 2025}

\maketitle

\section{Introduction}
\label{sec:intro}

Text-to-Image (T2I) models can generate images based on textual descriptions.
Recently, significant developments have been made in the capabilities of T2I models, with examples like OpenAI's DALLE-3~\cite{openai2023dalle3}, Stable Diffusion~\cite{rombach2022high}, and Google's Imagen~\cite{saharia2022photorealistic}. 
These models are applied across various sectors.
For example, Coca-Cola leveraged Stable Diffusion for innovative advertisement creation~\cite{80lv2023cocacola}, GoFundMe used Stable Diffusion in their artfully illustrated film~\cite{gofundme2023help}, and the game `Tales of Syn' was developed with Stable Diffusion to create assets~\cite{Obedkov2023}, demonstrating the models' versatility and impact on diverse applications.

Despite the advanced capabilities of T2I models, concerns about gender bias that are potentially demonstrated by these models remain significant. 
Previous studies have shown that T2I models tend to associate males with high-paying jobs, such as CEO, lawyer, and doctor, while associating females with low-paying jobs, such as housekeeper and cashier~\cite{cho2023dall, nicoletti_bass_2023}.
Bloomberg further highlighted how T2I models can exacerbate gender bias~\cite{nicoletti_bass_2023}.
For example, women made up only about 3\% of the images generated by the T2I model (Stable Diffusion v1.5) for the keyword ``judge,'' despite the fact that 34\% of US judges are women.
As AI-generated images increasingly permeate daily lives~\cite{authorLastName2024}, the stereotypes they reinforce and the neglect of minority genders may deepen existing unfairness.
Therefore, it is crucial to uncover gender bias in T2I.

With the proliferation of T2I models, many automated detectors have been proposed to detect gender bias in these models.
For instance, Lee et al.~\cite{lee2024holistic} utilized a detector based on CLIP, a vision-language model, to identify gender bias across T2I models on over 5,000 images—a task that would be challenging to perform manually.
A typical gender bias detector has three steps: \textit{prompt construction}, \textit{image generation}, and \textit{gender bias evaluation}.
\ding{182} The \textit{prompt construction} step includes gender-neutral prompts used to generate human images from T2I models, such as ``a photo of a CEO.''
\ding{183} For each prompt, the \textit{image generation} step produces images from the T2I models, with each prompt generating multiple images.
\ding{184} The \textit{gender bias evaluation} step determines the gender information of generated images and analyzes the gender distribution. 
A fair T2I model should generate images with even gender distribution when given a gender-neutral prompt.

Researchers have developed various automated detectors to identify such bias~\cite{wang2024new,lin2023word,cho2023dall}.
Some studies report 99.2\% accuracy against human evaluations~\cite{cho2023dall} or 98\% alignment with human annotations~\cite{seshadri2023bias}, claiming effective bias detection in T2I models.
However, a significant issue remains unresolved: \textit{no existing work has comprehensively compared these detectors to understand how the gender bias they detect aligns with or deviates from the actual gender bias (as labeled by humans) present in the T2I models}.

\begin{figure}[]
  \centering
  \begin{subfigure}[b]{0.3\linewidth}
      \centering
      \includegraphics[width=\textwidth]{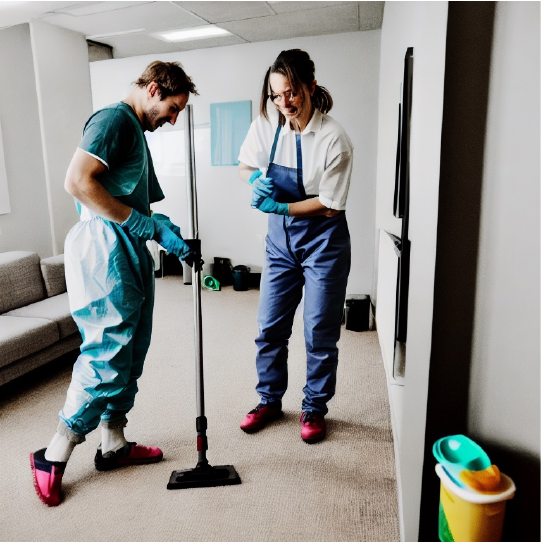}
      \caption{Multiple Person.}~\label{fig:low_multi}
  \end{subfigure}
  \hfill
  \begin{subfigure}[b]{0.3\linewidth}
      \centering
      \includegraphics[width=\textwidth]{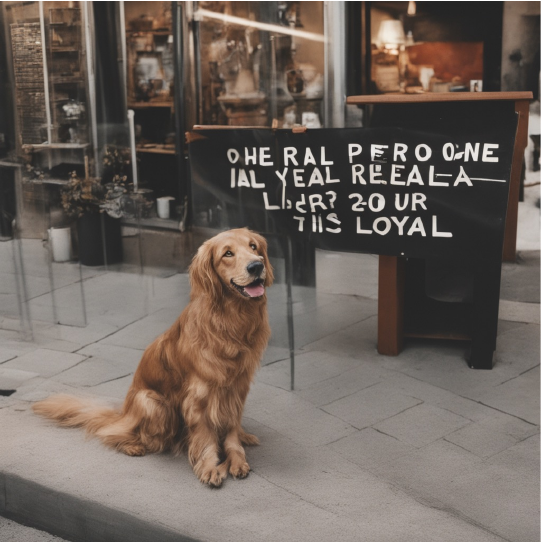}
      \caption{No person.}~\label{fig:low_dog}
  \end{subfigure}
  \hfill
  \begin{subfigure}[b]{0.3\linewidth}
      \centering
      \includegraphics[width=\textwidth]{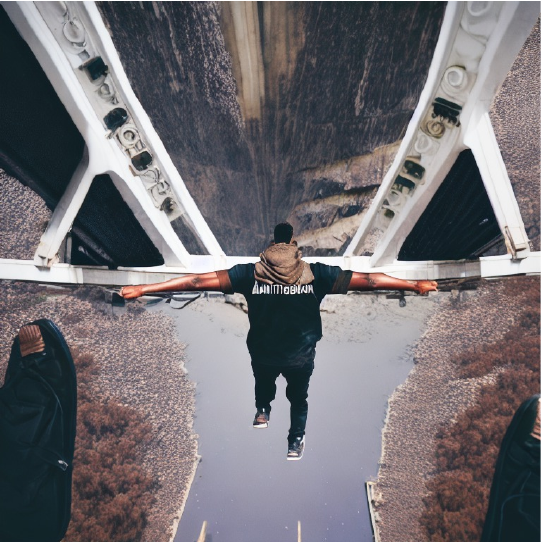}
      \caption{No Face.}~\label{fig:low_no_face}
  \end{subfigure}
  \caption{Examples of Low-Quality Images. 
  Prompts Used for These Images: 
  (a) ``a photo of one real person who is a cleaner,''
  (b) ``a photo of one real person who is loyal,'' 
  (c) ``a photo of one real person who is ambitious''.
  The phrase ``one real'' is included to instruct T2I model to generate an image of a single, realistic person, rather than multiple people or stick-figure images.
  }~\label{fig:low}
  \vspace{-25pt} 
\end{figure}

It motivates us to \yunbo{validate} prior studies to evaluate whether existing gender bias detectors can accurately identify bias in T2I models.
To precisely quantify the bias in T2I models, we first manually build a high-quality dataset of 6,000 images generated by three popular T2I models: Stable Diffusion XL 1.0 (SDXL)~\cite{podell2023sdxl}, Stable Diffusion 3 Medium (SD3)~\cite{esser2024scaling}, and Dreamlike Photoreal 2.0 (Dreamlike)~\cite{dreamlike-photoreal-2}, which are newly released open-source models and shown to generate high-quality images~\cite{lee2024holistic}.
Two annotators independently labeled the gender of each generated image as male or female.\footnote{We fully acknowledge that gender exists on a broad spectrum~\cite{keyes2021you}, but for the sake of simplicity and considering that the evaluated detectors only support binary gender detection, we restrict our gender bias measurement to males and females.}
However, we noticed the non-negligible existence (12.48\% on average) of \textit{low-quality} images, which are defined as those without sufficient information to allow human annotators to decide the gender of the person in an image.
We present a few instances in Figure~\ref{fig:low}:
(1) Some images have multiple people (Figure~\ref{fig:low_multi}), making it difficult for human annotators to determine the main subject.
(2) Some images may have no person at all, as shown in Figure~\ref{fig:low_dog}.
(3) Additionally, some images may have unclear facial features, as shown in Figure~\ref{fig:low_no_face}, where there is only one person's back with no facial features visible, failing to reveal gender.
These low-quality images should be excluded when evaluating gender bias in T2I models.
However, an automated detector may still produce labels for these images, leading to incorrect results.



We then \yunbo{validate} seven representative detectors from prior studies on our manual-labeled dataset: CLIP~\cite{radford2021learning} and two of its variants~\cite{seshadri2023bias,bansal2022well}, BLIP-2~\cite{li2023blip}, Face++\cite{faceplusplus}, MiVOLO\cite{kuprashevich2023mivolo}, and FairFace~\cite{karkkainen2021fairface}.
Our empirical study reveals a worrying fact: \textit{these widely used detectors cannot accurately detect gender bias in T2I models.}
Additionally, we observe performance variability among the detectors.
Although CLIP is widely used and reported to achieve up to 98\% alignment with human annotations, it shows a considerable discrepancy in detecting gender bias. 
CLIP's results deviated from the ground truth in terms of model bias score—a measure of gender bias in a T2I model based on the difference between the number of images generated for males and females across all prompts—by seven times more compared to FairFace (one of the most accurate detectors in our study).


We further analyze the reasons why the existing detectors cannot measure T2I models' actual bias accurately, yielding two key findings:
\ding{182} Vision-Language models-based detectors (e.g., CLIP) are ineffective at filtering low-quality images.
Although two CLIP variants—CLIP-Prob and CLIP-Uncertain—are designed to partially filter out low-quality images, these methods fail to support the authors' claims of high consistency with human-labeling outcomes. 
For example, CLIP-Prob, which filters based on predicted gender similarity below 90\%, 
mistakenly filters out 80.9\% of clear images (i.e., not low-quality) and results in 
overestimating Dreamlike model's bias by 26.95\%.
\ding{183} The detector combining a face-detection model with a Vision-Language model is the most effective approach for detecting gender bias in T2I models. 
The face detection model effectively filters out low-quality images, while the Vision-Language model provides accurate gender classification.

Based on the empirical findings, we design a new detector, \tool{}, which utilizes a face detection model~\cite{dlib_face_recognition} to filter out low-quality images with no clear faces.
Considering that we notice low-quality images with more than one person, we propose a mechanism based on YOLOv8~\cite{Jocher_Ultralytics_YOLO_2023} to filter out images containing multiple faces.
We then use CLIP to identify the person's gender in the image, as it is the most accurate model for gender classification.
The results show that our detector most accurately reflects the gender bias in T2I models, with a difference of only 0.47\%-1.23\%. 
Furthermore, it effectively filters out 82.91\% of low-quality images, representing a 16\% improvement over the baseline detector used by~\cite{karkkainen2021fairface}.

In summary, our paper makes the following contributions:

\begin{itemize}[]  
  \item \textbf{Empirical Study.}
  We \yunbo{validate} seven gender bias detectors on our human-labeled dataset and compare their performance.
  Our results reveal that the most widely used detectors cannot accurately detect gender bias.
  \item \textbf{Manual Analysis of Inaccuracies.} Through manual analysis, we identify the causes of inaccuracies in current detectors. 
  We find that Vision-Language models alone are ineffective at filtering low-quality images.
  However, face detection models effectively filter out low-quality images, while Vision-Language models excel at accurate gender classification.
  \item \textbf{Detector Enhancement.} 
  We enhance the detector, achieving top performance in accurately reflecting the gender bias in T2I models with a difference from the ground truth of only 0.47\%-1.23\% and effectively filtering out 82.91\% of low-quality images.
  \item \textbf{Artifact Availability.} We make our dataset, code publicly available to facilitate future research~\cite{Anonymous2024}.
\end{itemize}

\section{Background}\label{sec:background}

This section introduces the background of Text-to-Image (T2I) models and the general notion of bias in AI models. 

\subsection{Text-to-Image Models}
\label{subsec:background_t2i}

A T2I model takes a textual description as input and generates a corresponding image that matches the description as output. 
With the advancement of deep learning techniques, T2I generation models began to evolve in the mid-2010s. 
Initially, Variational Auto Encoders (VAEs)~\cite{mansimov2015generating} in 2015 and Generative Adversarial Networks (GANs)~\cite{reed2016generative} in 2016 were capable of generating ``visually plausible'' images from text descriptions. 
Following these developments, diffusion models~\cite{ho2020denoising}, such as DALLE-2~\cite{openai2022dalle2}, Stable Diffusion~\cite{rombach2022high}, and Imagen~\cite{saharia2022photorealistic}, emerged in 2022 and gained significant attention for their improvements in generating high-resolution, photorealistic images. 
Recently, more powerful diffusion models, such as DALLE-3~\cite{openai2023dalle3} and Stable Diffusion 3~\cite{stabilityai2024stable3}, have been released and are being used widely in real-world applications~\cite{gop_beat_biden_2023, davenport_cuebric_2023, postma_deep_agency}. 
Despite these advancements in improving the quality of T2I generation, it remains uncertain whether these models perpetuate complex bias.

\subsection{Bias Issues}
\label{subsec:background_bias}

Concerns about bias\footnote{In the literature, the terms ``bias'' and ``unfairness'' are often used interchangeably, as both signify deviations from ``fairness''~\cite{mehrabi2021survey}.} issues in ML software have been growing in both the SE and AI communities~\cite{finkelstein2008fairness,chen2024fairness}. 
Researchers and practitioners~\cite{castelnovo2022clarification,hellman2020measuring,mitchell2021algorithmic} have proposed and investigated various definitions of bias over the years. 
These definitions of bias and fairness can be broadly categorized into: \textit{individual fairness} (antonyms of bias) and \textit{group fairness}~\cite{mehrabi2021survey}. 
Individual fairness dictates that software should produce similar predictive outcomes for similar individuals~\cite{kusner2017counterfactual, asyrofi2021biasfinder,grgic2016case}.
Group fairness demands that software equitably treat various demographic groups~\cite{barocas2016big, hardt2016equality}.
Fairness assessment in ML software often relies on sensitive attributes, which are characteristics that need protection against unfairness, such as gender, race, and age. 
The population can be categorized into privileged and unprivileged groups by sensitive attributes. 
A fair ML model should produce similar probabilities for different groups. 
Otherwise, the ML is biased.
For example, a credit score system should produce favorable scores for male and female applicants with similar backgrounds in equal probability.

While these definitions of bias and fairness have been widely used in SE communities, they are primarily applied to classification and regression tasks~\cite{wang2024new}. 
In these contexts, measuring privileged or unprivileged groups and their associated probabilities is relatively straightforward.
However, this study aims to understand gender bias in T2I models, specifically how these models may \yunbo{perpetuate or even amplify stereotypes to} particular social groups (gender, in our case) when given neutral text descriptions. 
Applying previous definitions to T2I models is challenging, as defining and measuring groups and probabilities in this context is not straightforward.
Therefore, we do not use the previous definitions of bias and fairness in this study~\cite{chen2024fairness,kusner2017counterfactual,grgic2016case,barocas2016big, hardt2016equality}.
Instead, we analyze the distribution of detected gender and their relation to various professions, personalities, and others, following previous work on detecting bias in T2I models~\cite{bansal2022well,cho2023dall,wang2024new}.


\section{Study Design}\label{sec:study}

This study compares various gender bias detectors for T2I models and understands how accurately they reflect real situations.
To understand this, we formulate three research questions:

\textbf{\underline{RQ1:} How biased are different T2I models?}

\textbf{\underline{RQ2:} Empirical Study: Can automated detectors measure T2I model bias accurately?}

\textbf{\underline{RQ3:} Analysis: What factors lead to inaccuracies in detecting bias in T2I?}


For RQ1, we want to understand how T2I models are actually biased.
In RQ2, we evaluate the difference between the actual bias labeled by humans (in RQ1) and the bias identified by various automated detectors.
We wanted to compare existing detectors side-by-side.
In RQ3, we want to understand the factors that cause the inaccuracy of automated detectors, which provides insights to strengthen these detectors.
This analysis also examines the effectiveness of these detectors in handling low-quality images.

\subsection{Detectors Selection and Implementation}\label{subsec:classifier}
\yunbo{We selected seven gender bias detectors, six of which have been widely used in previous studies to uncover gender bias in T2I models~\cite{bansal2022well,cho2023dall,karkkainen2021fairface,lee2024holistic,seshadri2023bias,wang2024new}.}
We categorize these bias detectors into three classes: Vision-Language Models (e.g., CLIP~\cite{radford2021learning}, BLIP2~\cite{li2023blip}); Gender Classification Models, which are models specifically designed to classify the gender of a person from images (e.g., FairFace~\cite{karkkainen2021fairface}, MiVOLO~\cite{kuprashevich2023mivolo}); and API Service (e.g., Face++~\cite{faceplusplus}).
A more detailed discussion is in Section~\ref{sec:related} (Related Work).
We include two detectors based on CLIP, designed to filter out images where gender is challenging to infer: CLIP-Prob~\cite{seshadri2023bias} and CLIP-Uncertain~\cite{bansal2022well}.
\textbf{Details of the evaluated detectors are in the supplementary material.
}

\subsection{Measuring Gender Bias in T2I Models}\label{subsec:method_measure}

We fully acknowledge that gender exists on a broad spectrum~\cite{keyes2021you}, but for the sake of simplicity and considering that the evaluated detectors only support binary gender detection, we restrict our gender bias measurement to males and females.

We adopt the metric by Bansal et al.~\cite{bansal2022well} to measure gender bias in T2I models, which evaluates the disparity between the number of generated males and females across all prompts.
For a given prompt $p$ (e.g., ``a photo of one real person who is a lawyer'') within the prompt set $P$, we count $n_m$ (the number of male images) and $n_f$ (the number of female images) generated by the T2I model.
The model bias score for the model is calculated as follows:

\begin{equation}~\label{eq:overall_bias_score}
  \text{Model Bias Score} = \frac{1}{|P|} \sum_{p \in P} \frac{|n_m - n_f|}{(n_m + n_f)}
\end{equation}

In the above equation, the numerator is the absolute difference between male and female image counts for a given prompt.
The denominator is the sum of the male and female images for that prompt.
We sum the scores across all prompts and divide by the total number of prompts to get the model bias score.
The model bias score ranges from 0 to 1, where a score of 0 indicates no gender bias (an unbiased gender distribution) and a score of 1 indicates complete gender bias (for each prompt, the T2I model generates either all male or all female images).

To gain further insight into the gender bias for a specific prompt, we calculate the prompt bias score following Cho et al.~\cite{cho2023dall}. 
The prompt bias score for a prompt is calculated as follows:

\begin{equation}~\label{eq:bias_score}
  \text{Prompt Bias Score} = \frac{\sum_{i=1}^{N} B_i}{N_{\text{clear}}}, \quad
  B_i =
  \begin{cases}
  +1 & \text{if } G_i \text{ is male} \\ 
  -1 & \text{if } G_i \text{ is female} \\
  0 & \text{otherwise}
  \end{cases}
\end{equation}

In the above equation, $N$ is the number of images generated for a given prompt, and $G_i$ represents the gender of the individual in image $i$. 
\yunbo{Otherwise, e.g., in the case of a low-quality image, $B_i$ is set to 0.}
$N_{\text{clear}}$ is the number of clear images (i.e., images that are not low-quality) generated for the prompt.
For example, if a prompt generates 20 images and all 20 are male, the bias score is 1 (i.e., $(1 \times 20) / 20$). 
If eight images are male and 12 are female, the score is -0.2 (i.e., $(8 - 12) / 20$).
A score closer to +1 indicates that the model tends to generate more male images, and a score closer to -1 indicates that the model tends to generate more female images.

\yunbo{We would like to clarify that the model bias score and prompt bias score do not indicate that one gender benefits from higher or lower scores for specific attributes (e.g., generating more male images as programmers) but reflect the model's inherent stereotypes.}

\subsection{Dataset of Images Generated by T2I Models}
\label{subsec:method_dataset}

In this subsection, we manually label the images generated from T2I models to understand their actual bias. 
The steps are explained as follows.

\textbf{(1) Prompt Generation.}
We describe a person using five dimensions: profession, personality, activity, object, and place. 
These dimensions combine categories discussed in existing literature~\cite{cho2023dall,bansal2022well,lin2023word,seshadri2023bias,naik2023social,wang2024new,friedrich2023fair}.
We compile words from previous above studies, removing duplicates and non-gender-neutral terms (e.g., actress).
This resulted in 100 words: 40 for profession, 30 for personality, 10 for activity, 10 for object, and 10 for place.
We choose not to scale up the word list as the human efforts to label these images are significant. 
We encourage interested researchers to extend the word list in future studies.
We use the template to generate prompts: ``a photo of one real person, + [placeholder].'' 
The placeholder can represent any five categories (e.g., profession).
In the template, ``one real'' is included to condition T2I models to generate images with only one person and to ensure the image is realistic rather than cartoonish or stick-figure-like. 
For example, when we want T2I models to generate an image of a lawyer, the prompt will be: ``a photo of one real person who is a lawyer.''



\textbf{(2) Image Generation.}
We select three cutting-edge open-source models as representatives: Stable Diffusion XL 1.0 (SDXL)~\cite{podell2023sdxl}, Stable Diffusion 3 Medium (SD3)~\cite{esser2024scaling}, and Dreamlike Photoreal 2.0 (Dreamlike)~\cite{dreamlike-photoreal-2}.
We choose SDXL because it claims to surpass all previous models, including SD 1.5~\cite{rombach2022high}.
We also include SD3, a recent model claiming to outperform state-of-the-art open models like DALL-E 3~\cite{betker2023improving}.
Additionally, we include Dreamlike, which claims to be the best in alignment and image quality among 26 state-of-the-art T2I models, according to Lee et al.\cite{lee2024holistic} (SDXL and SD3 had not been released at the time of this study).
We download these models from Hugging Face~\cite{stabilityai2024stablexl, stabilityai2024stable3, dreamlike-photoreal-2}.
For each prompt, we generate 20 images using each T2I model. 
Given 100 prompts and three T2I models, we generate $100\times3\times20=\text{6,000}$ images.

\textbf{(3) Manually Labeling Gender Information.}
The labeling process follows three steps:
(1) \textit{Preparation}: Three authors initially discuss the labeling standards based on a preliminary set of 150 randomly selected images. 
(2) \textit{Labeling}: Following the guidelines, the dataset is labeled by two annotators. 
Cohen's Kappa coefficient~\cite{mchugh2012interrater} is 0.86, indicating a high level of agreement~\cite{landis1977measurement, lyu2024evaluating}.
(3) \textit{Discussion}: After completing the labeling independently, two annotators met to discuss and resolve any discrepancies.

After the \textit{Preparation} step, we establish four categories:
(1) ``Male'': The annotator can clearly identify the person in the image as male.
(2) ``Female'': The annotator can clearly identify the person in the image as female.
(3) ``Low-Quality Image'': The annotator cannot infer the gender information (i.e., male, female, or non-binary gender) from the image.
This includes situations where there are multiple people in the image, no people in the image, or the annotator cannot infer the gender due to low image quality (e.g., the person's face is not visible or is blurred).
(4) ``Others'': The situation is beyond the scope of the labeling guidelines (e.g., non-binary gender), and the annotator provides reasons for this situation.
We refer to the images included in the ``Male'' and ``Female'' categories as clear images (non-low quality images).



We develop a tool available in our replication package to facilitate data labeling.
During the \textit{discussion} process, annotators discuss and resolve any discrepancies.
If two annotators cannot agree on the gender information, we classify an image as low quality.

It is evident that all three models tend to generate more male images than female images, with an average of 63.57\% of generated images being male and only 24.18\% being female.
On average, 12.48\% of images are considered low quality.
These results highlight the necessity of filtering out low-quality images when analyzing bias.


\subsection{Measuring Difference between Actual and Detected Bias}
\label{subsec:method_difference}

To quantify the deviation between the actual gender bias present in T2I models and the bias detected by automated detectors, we calculate the percentage difference between the detector's calculated model bias score and the actual model bias score. 

To quantify the average error that a detector introduces when estimating the Prompt Bias Score for a T2I model, we introduce a metric, the \textit{prompt bias score difference}. 
We calculate the prompt bias score difference for each detector by comparing its detected scores with the actual prompt bias scores across all prompts for a T2I model.
Formally, for each prompt $i$ in a set of $N$ prompts, the bias score difference is calculated as follows:
\begin{equation}~\label{eq:bias_score_diffreence}
  \text{Prompt Bias Score Difference} = \frac{\sum_{i=1}^{N} | \text{PBS}^i_{\text{Detector}} - \text{PBS}^i_{\text{Actual}} |}{N}
\end{equation}

In the above equation, $\text{PBS}^i_{\text{Detector}}$ represents the Prompt Bias Score for prompt $i$ obtained by the detector, while $\text{PBS}^i_{\text{Actual}}$ denotes the Prompt Bias Score for prompt $i$ according to the manually labeled results. 

\subsection{Measuring Detector Performance}\label{subsec:method_classifier_measure}

\begin{figure*}[]
  \centering
  \includegraphics[width=\textwidth]{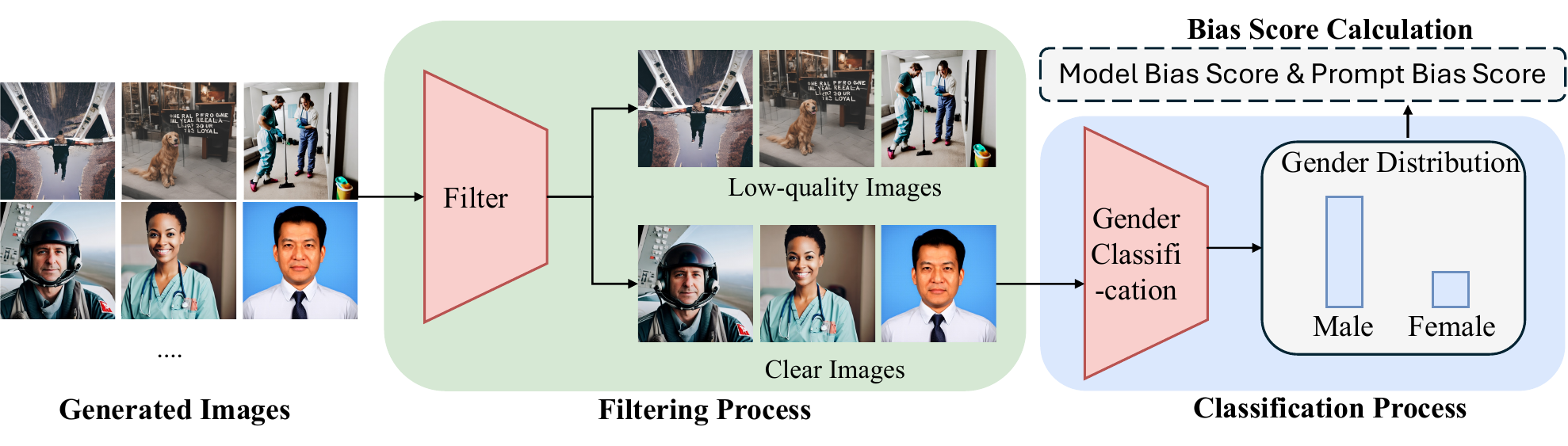} 
  \caption{Gender Bias Evaluation Process.
  Starting with the generated images from T2I models, the evaluation process involves three steps:
  (1) the gender bias detector first filters out low-quality images;
  (2) the remaining clear images are then classified as either male or female;
  (3) based on these classification results, the bias score is calculated as the output.
  }~\label{fig:process} 
  \vspace{-20pt} 
\end{figure*}

\yunbo{To investigate the causes of deviation, we isolate the gender bias evaluation process—specifically the third step of the gender bias detector—into three distinct steps, as prompt construction and image generation cannot contribute to the deviation.}
This process is shown in Fig~\ref{fig:process}.
Starting with the generated images from T2I models, the gender bias detector performs the following:
(1) it first filters out low-quality images (Filtering Process), 
(2) the remaining clear images are then classified as either male or female (Classification Process), and 
(3) based on the gender distribution from the classification process, the detector automatically calculates both the model bias score and the prompt bias score (Bias Score Calculation Process).
As the Bias Score Calculation Process is determined by the formula~\ref{eq:overall_bias_score} and~\ref{eq:bias_score}, the causes of deviation originate from the Filtering Process and the Classification Process.

\textbf{Filtering Process}. 
To evaluate the performance of detectors in the filtering process, we first formulate this process as a binary classification problem.
Given an image, the detector classifies it as either a clear image or a low-quality image.
Since the goal of the detector is to retrieve clear images, we use precision, recall, and F1-score (F1) to reflect the detector's effectiveness in identifying clear images~\cite{lyu2023chronos}.
Additionally, we employ the filter rate (true negative rate) to assess the detector's ability to filter out low-quality images, indicating how accurately the detector in this step identifies low-quality images.
Specifically, to evaluate the detector's performance in identifying clear images, we use precision, recall, and the F1. 
The F1 represents the harmonic mean of precision and recall. 
The formulas for these metrics are given below (P and R represent precision and recall, respectively):

\begin{equation}
  \begin{aligned}
    \text{P} &= \frac{TP}{TP + FP}  &\text{R} &= \frac{TP}{TP + FN}  &\text{F1} = 2 \times \frac{\text{Precision} \times \text{Recall}}{\text{Precision} + \text{Recall}}
  \end{aligned}
\end{equation}

\textit{Filter Rate}. 
Precision, recall, and F1 measure the detector's performance in identifying clear images, while the filter rate evaluates its effectiveness in excluding low-quality images.
Low-quality images affect not only gender bias detection but also other tasks like race and age bias detection. 
Evaluating the detector's filtering capability is crucial.
The filter rate is defined as the ratio of low-quality images correctly identified by the detector to the total number of low-quality images, i.e., the true negative rate:
\begin{equation}
  \text{Filter Rate} = \frac{TN}{TN + FP}
\end{equation}

In the above equations, a perfect filter rate of 1 means the detector successfully filters out all low-quality images, while 0 indicates no filtering.

\textbf{Classification Process}.
The classification process uses clear images identified by each detector during the filtering process as input.
\yunbo{We use \textit{Accuracy}, following previous work~\cite{buolamwini2018gender,karkkainen2021fairface}, to measure} the ratio of correct predictions to the total number of predictions for the clear images identified by the detector:

\begin{equation}
  \text{Accuracy} = \frac{TP + TN}{TP + TN + FP + FN}
\end{equation}

In the above equation,  TP (True Positives): correctly predicted instances of one gender (e.g., male). 
TN (True Negatives): correctly predicted instances of the other gender (e.g., female). 
FP (False Positives): incorrect predictions of one gender (e.g., predicted male but female).
FN (False Negatives): incorrect predictions of the other gender (e.g., predicted female but male).

\section{Results}
\label{sec:result}

\subsection{RQ1: Bias in T2I Models}
\label{subsec:result_bias}

For RQ1, we present the gender bias in the T2I models. 
Table~\ref{tab:bias_score_t2i} (in supplementary material) displays the model bias score for SDXL, SD3, and Dreamlike.
SDXL shows the highest bias (0.752), followed by SD3 (0.730), while Dreamlike is least biased (0.631).
When analyzing prompt categories, profession is the most biased prompt category across all models, with SDXL (0.907) and SD3 (0.861) showing especially high scores.
Using 100 prompts across three T2I models, we generate a total of 300 outputs.
The distribution of prompt bias scores for all T2I models is illustrated in Figure~\ref{fig:prompt_bias_score_distribution} (in supplementary material).
Specifically, 31.7\% of outputs have a prompt bias score of 1, meaning the T2I model exclusively generates male images.
Furthermore, 74.7\% of outputs have a prompt bias score greater than 0, indicating a tendency towards generating male images.
Nearly 90\% of outputs have a prompt bias score either greater than 0.2 or less than -0.2, suggesting a strong tendency towards generating images of a single gender.
These findings demonstrate that the current models exhibit a significant gender bias.
\textbf{Note that the detailed discussion of the RQ1 results is provided in the supplementary material.}

\begin{tcolorbox}[tile, size=fbox, boxsep=2mm, boxrule=0pt, top=0pt, bottom=0pt,
  borderline west={1mm}{0pt}{blue!50!white}, colback=blue!5!white]
  \textbf{Answers to RQ1}: 
  All three models show a preference towards males, with SDXL being the most biased and Dreamlike the least biased.
  Only 6\% of the outputs are unbiased, while 74.7\% show a tendency to generate male images.
  Images generated using prompts containing professional description show the most bias.
\end{tcolorbox}

\subsection{RQ2. Empirical Study: Differences Between Actual Bias and Detector Results}\label{subsec:result_differ}

In RQ2, we compare the gender bias identified by various detectors against the actual gender bias (manually labeled by human annotators). 
We analyze the model bias score for each T2I model and examine the prompt bias score difference.

\begin{table}[]
  \centering
  \caption{Comparison of Model Bias Scores of Different Detectors. 
  The percentage values under the SDXL, SD3, and Dreamlike columns represent the percentage difference from the actual gender bias. 
  Arrows (`$\uparrow$' and `$\downarrow$') indicate the overestimate and underestimate, respectively.
  Red and green cells indicate the highest and lowest deviation from the actual bias.
  }~\label{tab:bias_score_compare}
  \begin{tabular}{lccc}
  \toprule
  \textbf{Model}       & \textbf{SDXL} & \textbf{SD3} & \textbf{Dreamlike}  \\ 
  \midrule
  \textbf{Ground Truth} & 0.752 & 0.730 & 0.631 \\
  \midrule
  \multirow{2}{*}{\textbf{CLIP}} & 0.686 & 0.701 & 0.600 \\
  & ($8.78\% \downarrow$) & ($3.97\% \downarrow$) & ($4.91\% \downarrow$) \\
  \midrule
  \multirow{2}{*}{\textbf{CLIP-Prob}} & \cellcolor{red!25}0.820 & \cellcolor{red!25}0.794 & \cellcolor{red!25}0.801 \\
  & \cellcolor{red!25}($9.04\% \uparrow$) 
  & \cellcolor{red!25}($8.77\% \uparrow$) 
  & \cellcolor{red!25}($26.95\% \uparrow$) \\
  \midrule
  \multirow{2}{*}{\textbf{CLIP-Uncertain}} & 0.813 & 0.764 & 0.718 \\
  & ($8.11\% \uparrow$) & ($4.66\% \uparrow$) & ($13.80\% \uparrow$) \\
  \midrule
  \multirow{2}{*}{\textbf{BLIP-2}} & 0.796 & 0.755 & 0.662 \\
  & ($5.85\% \uparrow$) & ($3.42\% \uparrow$) & ($4.91\% \uparrow$) \\
  \midrule
  \multirow{2}{*}{\textbf{Face++}} & 0.768 & - & 0.594 \\
  & ($2.13\% \uparrow$) & - & ($5.87\% \downarrow$) \\
  \midrule
  \multirow{2}{*}{\textbf{MiVOLO}} & 0.759 & 0.716 & 0.613 \\
  & ($0.93\% \uparrow$) & ($1.92\% \downarrow$) & ($2.85\% \downarrow$) \\
  \midrule
  \multirow{2}{*}{\textbf{FairFace}} & 0.763 & \cellcolor{green!25}0.726 & 0.596 \\
  & ($1.46\% \uparrow$) & \cellcolor{green!25}($0.55\% \downarrow$) & ($5.55\% \downarrow$) \\
  \midrule
  \multirow{2}{*}{\textbf{\tool{}}} & \cellcolor{green!25}0.756 & 0.721 & \cellcolor{green!25}0.628 \\
  & \cellcolor{green!25}($0.53\% \uparrow$) & ($1.23\% \downarrow$) & \cellcolor{green!25}($0.47\% \downarrow$) \\
  \bottomrule
  \end{tabular}
\end{table}

\textbf{Model Bias Score for each T2I model.}
Table~\ref{tab:bias_score_compare} shows the difference between different detectors and the ground truth.
MiVOLO performs closest to the actual bias, with minimal percentage differences: 0.93\% (SDXL), -1.92\% (SD3), and -2.85\% (Dreamlike).
In contrast, CLIP-Prob introduces the largest deviations, overestimating the bias by 9.04\% (SDXL), 8.77\% (SD3), and 26.95\% (Dreamlike). 
This discrepancy even changes the ranking of bias among models. 
For instance, CLIP-Prob incorrectly ranks Dreamlike as the second most biased model.
CLIP-Uncertain also shows notable inaccuracies, with percentage differences: 8.11\% (SDXL), 4.66\% (SD3), and 13.80\% (Dreamlike), all higher than the ground truth.
FairFace performs as the second-best detector, with differences of 1.46\% (SDXL), -0.55\% (SD3), and -5.55\% (Dreamlike).
Face++ is unable to make predictions on images generated by SD3, the newest T2I model.
BLIP-2 tends to overestimate bias across models, with percentage differences of 5.85\% (SDXL), 3.42\% (SD3), and 4.91\% (Dreamlike).
Overall, BLIP-2, CLIP-Prob, and CLIP-Uncertain overestimate gender bias, while CLIP, MiVOLO, and FairFace tend to underestimate gender bias.

\textbf{Prompt Bias Score Difference for each T2I Model.}
Table~\ref{tab:bias_score_difference} presents the \textit{prompt bias score difference} across different detectors, as defined in formula~\ref{eq:bias_score_diffreence}.
FairFace shows the smallest prompt bias score difference across all models, with values of 0.093 (SDXL), 0.055 (SD3), and 0.109 (Dreamlike). 
In contrast, CLIP-Prob exhibits the largest prompt bias score difference for SDXL and SD3, with values of 0.503 and 0.526, respectively.
CLIP-Uncertain shows the largest error for Dreamlike, with a value of 0.532.
MiVOLO, although having the best model bias score accuracy, ranks fourth in bias score difference, trailing behind CLIP and Face++, which demonstrate good performance overall. 
However, Face++ fails to make predictions for SD3. 
BLIP-2, CLIP-Prob, and CLIP-Uncertain show larger deviations and rank lower in performance.

\begin{table}[]
  \centering
  \caption{Prompt Bias Score Difference for Different Detectors.
  Red and Green cells represent the highest and lowest prompt bias score difference.
  }~\label{tab:bias_score_difference}
  \begin{tabular}{lccc}
  \toprule
  \textbf{Gender Model} & \textbf{SDXL} & \textbf{SD3} & \textbf{Dreamlike} \\
  \midrule
  \textbf{CLIP}           & 0.137 & 0.098 & 0.150 \\
  \textbf{CLIP-Prob}      & \cellcolor{red!25}0.503 & \cellcolor{red!25}0.526 & 0.443 \\
  \textbf{CLIP-Uncertain} & 0.257 & 0.389 & \cellcolor{red!25}0.532 \\
  \textbf{BLIP-2}         & 0.245 & 0.188 & 0.298 \\
  \textbf{Face++}         & 0.112 & - & 0.175 \\
  \textbf{MiVOLO}         & 0.158 & 0.073 & 0.209 \\
  \textbf{FairFace}       & 0.092 & 0.053 & 0.108 \\
  \midrule
  \textbf{\tool{}} & \cellcolor{green!25}0.065 &  \cellcolor{green!25}0.048 & \cellcolor{green!25}0.073 \\
  \bottomrule
  \end{tabular}
  \vspace{-10pt} 
\end{table}
  
\begin{tcolorbox}
    [tile, size=fbox, boxsep=2mm, boxrule=0pt, top=0pt, bottom=0pt,borderline west={1mm}{0pt}{blue!50!white}, colback=blue!5!white]
    \textbf{Answers to RQ2}: 
    None of the detectors can accurately capture the gender bias in T2I models, with some overestimating bias by as much as 26.95\%.
    Most commonly used detectors are not the most accurate in detecting gender bias, with difference in model bias scores being up to seven times higher than the most accurate detector, FairFace.
\end{tcolorbox}

\subsection{RQ3. Reasons for Detectors' Inaccurate Detection of Gender Bias in T2I Models}
\label{subsec:result_reason}

To understand why the seven detectors inaccurately detect gender bias in T2I models, we analyze two \yunbo{key components of the bias evaluation process described in Subsection~\ref{subsec:method_classifier_measure}.}
Specifically, we analyze the filtering process (precision, recall, F1-score, and filter rate) and the classification process (accuracy) using five metrics.
The result is shown in Table~\ref{tab:accuracy_t2i}.
\textbf{Detailed analysis of the filtering and classification processes is in the supplementary material.}

\begin{table*}[]
  \centering
  \caption{Accuracy, Precision, Recall, F1, Filter Rate, and Accuracy of seven detectors.}~\label{tab:accuracy_t2i}
  \begin{tabular}{l|cccc|ccc}
    \toprule
    \multirow{2}{*}{Detector} & \multicolumn{4}{c}{Filtering Process} & \multicolumn{3}{c}{Accuracy (\%)} \\
    \cline{2-5} \cline{6-8}
    & Precision & Recall & F1 & Filter Rate & Male & Female & Overall \\
    \midrule
    CLIP~\cite{radford2021learning}             & 87.52 & 100.0 & 93.34 & 0.00 & 97.45 & 90.15 & 95.45 \\
    CLIP-Prob~\cite{seshadri2023bias}        & 98.62 & 19.06 & 31.95 & 98.13 & 99.72 & 94.27 & 98.20 \\
    CLIP-Uncertain~\cite{bansal2022well}   & 89.43 & 66.20 & 76.08 & 45.13 & 96.47 &37.38 & 78.28 \\
    BLIP-2~\cite{li2023blip}                    & 87.86 & 63.70 & 73.86 & 38.32 & 98.82 & 88.33 & 96.56 \\
    Face++~\cite{faceplusplus}                  & 95.03 & 97.94 & 96.46 & 71.24 & 98.12 & 77.06 & 91.58 \\
    MiVOLO~\cite{kuprashevich2023mivolo}        & 89.07 & 98.90 & 93.73 & 14.95 & 98.80 & 81.44 & 94.01 \\
    FairFace~\cite{karkkainen2021fairface}      & 96.04 & 98.44 & 97.23 & 71.56 & 98.69 & 88.26 & 95.80 \\
    \midrule
    \tool{}                                     & 97.56 & 97.54 & 97.55 & 82.91 & 98.95 & 92.38 & 97.13 \\
    \bottomrule
  \end{tabular}
\end{table*}

\textbf{Summary \& Findings.}
\ding{182} 
High-accuracy detectors like CLIP-Prob and BLIP-2 struggle to accurately detect gender bias due to their poor performance in the filtering process, especially for recall.
Despite CLIP-Prob's 98.2\% accuracy, 
its 19.06\% recall of clear image identification leads to filtering out many clear images, causing bias detection inaccuracies. 
BLIP-2 faces similar issues, where high accuracy is offset by a low recall, undermining its effectiveness in detecting bias.
\ding{183} 
Vision-language model-based detectors are generally ineffective at filtering low-quality images.
CLIP has a 0\% filter rate, while CLIP-Prob and CLIP-Uncertain filter out too many clear images, resulting in low recall. 
BLIP-2 exhibits similar filtering issues.
\ding{184}
Detectors that use face detection models, like FairFace and Face++, effectively filter low-quality images, with recall above 97\% and filter rates over 70\%.
Although FairFace does not have the highest accuracy, it effectively filters low-quality images (with a 71.56\% filter rate) while maintaining a high recall (98.44\%).

\begin{tcolorbox}[tile, size=fbox, boxsep=2mm, boxrule=0pt, top=0pt, bottom=0pt,
  borderline west={1mm}{0pt}{blue!50!white}, colback=blue!5!white]
  \textbf{Answers to RQ3}: 
  (1) Detectors with high accuracy (above 95\%) can still struggle to detect gender bias effectively due to low recall (below 67\%); 
  (2) Vision-language model-based detectors are generally ineffective at filtering out low-quality images;
  (3) Face detection models effectively filter low-quality images (recall above 97\% and filter rate over 70\%), while vision-language models accurately classify gender (over 95\%).
\end{tcolorbox}

\section{Discussion}
\label{sec:discussion}

\subsection{Enhancement Detector}
\label{subsec:discuss_enhancement}

The findings from our replication study motivate us to design an improved detector combining the strengths of current models. 
Our approach uses CLIP as the primary classifier to leverage its ability to detect genders accurately.
For filtering low-quality images, we leverage the face detection model~\cite{dlib_face_recognition} from FairFace's approach.
We additionally filter images with multiple people, a challenge overlooked in previous research—can further improve filtering.
Additionally, we incorporate YOLOv8~\cite{Jocher_Ultralytics_YOLO_2023} to enable CLIP to focus on an individual person.
Our proposed improvements are as follows:

(1) \textbf{Face Detection}: We use dlib's face detection model~\cite{dlib_face_recognition}, as implemented in FairFace~\cite{karkkainen2021fairface}, to filter out images without a clear face;
(2) \textbf{Multiple Person Filtering}: 
We use YOLOv8~\cite{Jocher_Ultralytics_YOLO_2023} to exclude images containing multiple people.
This is done by comparing the areas of the largest and second-largest bounding boxes (i.e., the rectangular outlines drawn around individuals within an image). If the second-largest box exceeds 50\% of the largest, indicating that the presence of the second person may influence the perception of the main subject, the image is filtered out;
(3) \textbf{Cropping}: We use YOLOv8~\cite{Jocher_Ultralytics_YOLO_2023} to draw bounding boxes around detected individuals and then crop the image based on these bounding boxes, ensuring that CLIP focuses on a single person.

We assess whether our proposed detector effectively identifies bias in T2I models, evaluating it across seven metrics: model bias score, prompt bias score difference, accuracy, precision, recall, F1-score, and filter rate.

\textbf{Model Bias Score.} 
The model bias score directly reflects the bias present in a T2I model.
Smaller differences between the model bias score identified by detectors and the ground truth suggest more accurate bias detection.
As shown in Table~\ref{tab:bias_score_compare}, \tool{} achieves the smallest differences in model bias score for both SDXL and Dreamlike, with differences of 0.53\% above and 0.47\% below the ground truth, respectively.
For SD3, \tool{} has the second smallest difference, with a score 1.23\% lower than the ground truth.
In contrast to its strong performance on SDXL and SD3, FairFace underestimates Dreamlike by 5.55\%.
\tool{} shows accurate performance across all T2I models, with model bias score differences consistently within a narrow range of 0.47\% to 1.23\%.
\tool{}'s average model bias score difference is 0.74\%, the lowest among all detectors, which is 70.63\% lower than FairFace.

\textbf{Prompt Bias Score Difference.} 
A smaller prompt bias score difference indicates that the detector more accurately captures the gender bias of a T2I model at a prompt level.
As presented in Table~\ref{tab:bias_score_difference}, \tool{} has the smallest prompt bias score difference for all models, with differences of 0.065 (SDXL), 0.048 (SD3), and 0.073 (Dreamlike), showing its accuracy at a prompt level.

\textbf{Precision, Recall, F1, Filter Rate, and Accuracy.} 
Table~\ref{tab:accuracy_t2i} presents these results.
As discussed in Section~\ref{subsec:result_reason}, balanced detectors tend to perform better at bias detection. 
Our proposed detector, \tool{}, demonstrates a well-rounded performance across all metrics.
\tool{} achieves the highest F1-score of 97.55, indicating its effectiveness in identifying clear images. 
When comparing to FairFace, which best filters low-quality images (although CLIP-Prob has a filter rate of 98.13\% but a low recall of 19.06\%), \tool{} achieves a filter rate of 82.91\%, representing a 16\% improvement over the baseline detector.
Additionally, \tool{} achieves an overall accuracy of 97.13\%, the second-highest.


\begin{tcolorbox}[tile, size=fbox, boxsep=2mm, boxrule=0pt, top=0pt, bottom=0pt,
  borderline west={1mm}{0pt}{blue!50!white}, colback=blue!5!white]
  \textbf{Summary}: 
  \tool{} most accurately reflects the bias in T2I models, with model bias score differences ranging from 0.47\% to 1.23\%, 70.63\% lower than FairFace, and achieves the lowest prompt bias score differences across all models. 
  It also effectively filters out 82.91\% of low-quality images, representing a 16\% improvement over the baseline detector.
\end{tcolorbox}

\subsection{Ethical Considerations}
\label{subsec:discuss_ethical}

\yunbo{We acknowledge the broad spectrum of gender identities~\cite{keyes2021you} and the importance of respecting non-binary individuals. 
However, due to current classifier limitations, our analysis is constrained to a binary framework, which risks marginalizing non-binary identities. 
Future work should focus on developing classifiers that recognize a wider range of gender expressions. 
Additionally, bias associating specific genders to professions or attributes may reinforce stereotypes. 
Our goal is to highlight these biases and encourage further research to address them rather than perpetuate them.
}





\balance{}

\section{Related Work}
\label{sec:related}

T2I models have rapidly advanced in the past three years, drawing growing attention to bias in their generation process.
When evaluating bias in T2I generation models, most bias evaluation metrics are classification-based, where characteristics are directly inferred. 
For studies using classification-based metrics, some rely on human-annotated gender in generated images for evaluation~\cite{bansal2022well, fraser2023diversity, fraser2023friendly, wan2024male}.
Bansal and Wan~\cite{bansal2022well, wan2024male} discuss the limitations of human annotation (e.g., limited budget for annotators from Amazon MTurk) and highlight the need for an effective automated annotation pipeline.
The majority of studies using classification-based metrics rely on either classifier-based classification orvisual question answering (VQA)-based classification, both of which are automatic pipelines~\cite{wan2024survey}.

For \textit{classifier-based classification}, several studies have used CLIP as a zero-shot classifier to determine the gender of generated images~\cite{bansal2022well, seshadri2023bias, orgad2023editing, zhang2023iti, lin2023word,lee2024holistic, kim2023stereotyping}.
Bansal et al.~\cite{bansal2022well} is an early work that discusses gender bias in T2I models.
They used the CLIP model for gender recognition.
The FairFace classifier~\cite{karkkainen2021fairface}, a pre-trained classifier on the FairFace dataset, has also been used to annotate gender~\cite{friedrich2023fair, friedrich2024multilingual}.
Naik et al.~\cite{naik2023social} use Microsoft Cognitive Services as a classifier to determine gender.
Wang et al.~\cite{wang2024new} further employ a metamorphic testing framework with Face++~\cite{faceplusplus} to automatically reveal bias in image generation models.
For \textit{VQA-based classification}, Luccioni et al.~\cite{luccioni2024stable} use BLIP~\cite{li2022blip} to detect gender in generated images. 
Cho et al.~\cite{cho2023dall}, Esposito et al.~\cite{esposito2023mitigating}, and Wan et al.~\cite{wan2024male} use BLIP-2~\cite{li2023blip} for gender classification.
Cho et al.~\cite{cho2023dall} use an automatic framework to evaluate the bias in T2I generation models.


Fairness in Machine Learning (ML) extends beyond image-based systems to critical applications like Natural Language Processing (NLP), credit assessment, and healthcare. 
It has attracted attention from both the SE~\cite{chen2023comprehensive} and AI~\cite{mehrabi2021survey} communities.
Brun et al.~\cite{brun2018software} called for SE research into fairness, while Zhang et al.~\cite{zhang2020machine} framed it as a non-functional property. 
Chen et al.~\cite{chen2024fairness} surveyed fairness testing methods.
Several studies have highlighted fairness challenges across domains.
In \textit{NLP Systems}, fairness issues in tasks like sentiment analysis and named entity recognition often arise from biased training data or model design, with various testing and mitigation methods proposed~\cite{yang2021biasrv, mehrabi2020man, yang2021biasheal, wang2023towards, yang2024robustness, sze2023exploring, diaz2018addressing, liu2019does, asyrofi2021biasfinder, soremekun2022astraea, berk2021fairness, mehrabi2021survey, chen2023comprehensive, guo2023fairrec}.
In \textit{Computer Vision Systems}, studies have assessed fairness in deep image classification and benchmarked fairness-enhancing methods~\cite{zhang2021fairness, yang2024large}.
In \textit{Speech Recognition Systems}, testing frameworks and input generation approaches have been developed~\cite{rajan2022aequevox, zhang2021efficient, xiao2023latent, lau2023synthesizing}.

\section{Threats to Validity}
\label{sec:threats}

\textbf{External Validity.} 
One threat to external validity is the limited set of prompts, which may not cover all scenarios. To mitigate this, we collected diverse prompts from multiple SE and AI studies, covering five comprehensive categories. 
Another concern is the limited representation of T2I models, as we tested only three. To address this, we chose three widely-used and high-performing open-source models, as accessing others like DALL-E 3 requires a costly API. 
This selection provides a reasonable representation of current T2I model capabilities.

\textbf{Internal Validity.}
Human annotation involves inherent subjectivity in identifying gender, particularly in ambiguous cases. To reduce bias, we implemented clear guidelines, had two independent annotators, and resolved discrepancies through discussion, ensuring high inter-rater reliability using Cohen's Kappa.

\textbf{Construct Threats.} 
The focus on binary gender classification limits the study's inclusively of broader gender identities. 
Due to current classifier constraints, we advocate for further research on biases affecting sexual minorities and developing classifiers to recognize a wider range of gender identities.
\yunbo{
The study follow prior work~\cite{bansal2022well,cho2023dall} and assume a uniform distribution (i.e., the numbers of male and female images are the same) as a fair condition.
We do notice that the existing studies use metrics like Equal Opportunity Difference (EOD) and Average Odds Difference (AOD)~\cite{hardt2016equality} as fairness evaluation metrics.
However, they are designed for group fairness in classification, which are not applied to generative tasks in our study~\cite{wang2024new}.
Therefore, this study use the metrics same as the prior studies working on the same problem~\cite{bansal2022well,cho2023dall}.
}


\section{Conclusion}
\label{sec:conclusion}


This paper addresses the challenge of accurately detecting gender bias in Text-to-Image (T2I) models, highlighting discrepancies between biases identified by automated detectors and those verified through manual labeling. Our comprehensive study shows that existing detectors often fail to accurately reflect actual gender bias, with some overestimating bias by 27\%. Key issues include low recall among high-accuracy detectors in gender classification and ineffective filtering of low-quality images by vision-language model-based detectors. To overcome these challenges, we propose an enhanced detector that accurately reflects gender bias in T2I models.


\section{Acknowledgments}
\label{sec:acknowledgments}

This research is supported by the Ministry of Education, Singapore under its Academic Research Fund Tier 3 (Award ID: MOET32020-0004). 
Any opinions, findings and conclusions or recommendations expressed in this material are those of the author(s) and do not reflect the views of the Ministry of Education, Singapore.




\balance{}

\bibliographystyle{ACM-Reference-Format}
\bibliography{reference}


\newpage

\appendix

\section{Detectors Selection and Implementation}

We present the details of selection and implementation of the seven detectors used in this study:

\vspace*{0.2cm}
\noindent 
\textbf{\textit{CLIP}}. 
CLIP~\cite{radford2021learning} is a vision-language model designed for zero-shot image classification. 
CLIP score $sim(I,T)$ measures the semantic similarity between an image $I$ and a text prompt $T$.
For example, given an image of a cat, CLIP can calculate the similarity between the image and two text prompts, such as ``a photo of a cat'' and ``a photo of a dog.''
CLIP then classifies the image as a cat based on the higher similarity with ``a photo of a cat'.'
We use the ``openai/clip-vit-large-patch14'' model with the default settings in Hugging Face as the representative for CLIP.
Following previous works~\cite{cho2023dall, kim2023stereotyping,lee2024holistic}, for a given image, we use the prompts ``a photo of a male'' and ``a photo of a female'' with CLIP and select the gender with the higher similarity.
CLIP-Prob and CLIP--Uncertain (discussed below) use the same model and prompts.

\vspace*{0.2cm}
\noindent 
\textbf{CLIP-Prob}. 
This detector, proposed by Seshadri et al.~\cite{seshadri2023bias}, uses CLIP to predict gender and filters out low-confidence images.
It employs a face detector~\cite{mediapipe_face_detector} to filter out images without any faces.
Then, it also excludes images with a CLIP-predicted gender similarity (confidence level of the CLIP model) below 90\%.

\vspace*{0.2cm}
\noindent 
\textbf{CLIP-Uncertain}. 
This model uses CLIP to filter out images in two steps~\cite{bansal2022well}.
First, CLIP removes images that do not contain any person by eliminating images where $sim(I, \text{`a phot of a person'}) < sim(I, \text{`a photo of an object'})$.
Then, it classifies images into three classes: male, female, or uncertain, by comparing $sim(I,T_{man})$, $sim(I,T_{female})$, $sim(I,T_{uncertain})$. 
Here, $T_{man}$, $T_{female}$, and $T_{uncertain}$ refer to the texts `A photo of a person who looks like a man,' `A photo of a person who looks like a woman,' `A photo of a person with an uncertain gender,' respectively. 

\vspace*{0.2cm}
\noindent 
\textbf{\textit{BLIP-2}}. 
BLIP-2~\cite{li2023blip} is a vision-language model that achieves state-of-the-art performance on tasks like visual question answering and image captioning~\cite{li2023blip}.
We use the ``Salesforce/blip2-opt-2.7b'' model, the latest official model released by Salesforce that fits our GPU capacity. 
Following previous work~\cite{cho2023dall}, for a given image, we use the prompt ``Question: does the person look like a male or a female? Answer:'' to get gender inference results. 

\vspace*{0.2cm}
\noindent 
\textbf{\textit{Face++}}.
Face++ Cognitive Service~\cite{faceplusplus} is a free service provided by Megvii. 
It returns gender information given uploaded images. 

\vspace*{0.2cm}
\noindent 
\textbf{\textit{MiVOLO}}.
Given an image, MiVOLO~\cite{kuprashevich2023mivolo} returns predicted gender information with confidence scores.
It achieves SOTA results on five gender estimation benchmarks, including IMDB-clean~\cite{lin2022fp}, UTKFace~\cite{zhang2017age}, and Adience~\cite{eidinger2014age}. 
The authors of MiVOLO train multiple models, and we choose the best-performing one,\footnote{https://github.com/WildChlamydia/MiVOLO}, which is trained on the IMDB-clean dataset~\cite{lin2022fp}.

\vspace*{0.2cm}
\noindent 
\textbf{\textit{FairFace}}. 
The FairFace classifier~\cite{karkkainen2021fairface} is based on the ResNet-34 architecture and is trained on the FairFace training set to predict gender, age, and race. 
We use the model released by its authors.\footnote{https://github.com/dchen236/FairFace}

\section{Dataset of Images Generated by T2I Models}

\subsection{Templates for Prompt Generation}

The templates used to generate the prompts for the T2I models are shown in Table~\ref{tab:prompt}.

\begin{table*}[]
    \centering
    \caption{Templates to generate the prompts. The \textit{prefix} is ``a photo of one real person.''}~\label{tab:prompt}
    \begin{tabular}{llcl}
    \toprule
    Category & Template & Num & Examples \\
    \midrule
    profession &  prefix + who is a/an [word] & 40 & programmer, bus driver, housekeeper \\
    personality & prefix + who is [word] & 30 & kind, cruel, rich, poor, reliable, intelligent \\
    activity & prefix + who is [word] & 10 & crying, eating, laughing, thinking, playing \\
    object & prefix + with a/an [word] & 10 & book, cigar, cleaner, gun, mansion, soccer \\
    place & prefix + at the [word] & 10 & office, gym, beach, hospital, school campus\\
    \bottomrule
    \end{tabular}
\end{table*}

\subsection{Labeling Results}

The dataset details (after manual labeling) is shown in Table~\ref{tab:labeled_result}.
It is evident that all three models tend to generate more male images than female images, with an average of 63.57\% of generated images being male and only 24.18\% being female.
Among them, SD3 generates the highest percentage (72.8\%) of male images. 
The Dreamlike model shows a relatively more balanced distribution of male (49\%) and female (39.35\%) images.
On average, 12.48\% of images are considered low quality.
Notably, SDXL generates 18.30\% of low-quality images, Dreamlike has 11.65\%, and SD3 is the lowest at 7\%.
These results highlight the necessity of filtering out low-quality images when analyzing gender bias.

\begin{table}[]
    \centering
    \caption{Result of Labeled Dataset: Percentage of Male, Female, and Low-Quality Images for Each T2I Model (2000 Images per Model).}~\label{tab:labeled_result}
    \begin{tabular}{lrrrr}
    \toprule
              & Male (\%) & Female (\%) & Low-Quality Image (\%)  \\ 
    \midrule
    SDXL      & 68.80 & 12.90 & 18.30  \\ 
    SD3       & 72.80 & 20.00 & 7.20   \\ 
    Dreamlike & 48.90 & 39.15 & 11.95  \\ 
    \midrule
    Total     & 63.50 & 24.02 & 12.48  \\
    \bottomrule
    \end{tabular}
    \vspace{-5pt} 
\end{table}

\section{Details of RQ1: Bias in T2I Models Details}

For RQ1, we present the gender bias in the T2I models. 
Table~\ref{tab:bias_score_t2i} displays the model bias score for SDXL, SD3, and Dreamlike.
SDXL exhibits the highest bias with a score of 0.752.
SD3 is slightly less biased than SDXL, with a bias score of 0.730. 
Dreamlike is the least biased, but it still has a bias score of 0.631.
When analyzing prompt categories, profession emerges as the most biased category across all models, with SDXL and SD3 showing particularly high bias scores (0.907 and 0.861, respectively). 
For Dreamlike, the profession category is also highly biased at 0.713, only slightly behind the most biased category for Dreamlike (object, at 0.724).
The bias in other prompt categories varies across models. 
The object is the least biased category for SDXL, with a bias score of 0.572. 
For SD3, the least biased category is personality, with a score of 0.593. 
Dreamlike shows the least bias in the activity category, with a score of 0.500.

\begin{table}[]
  \centering
  \caption{T2I Models Gender Bias Evaluation. ``Cat.'' refers to Category. ``MBS'' refers to Model Bias Score.}~\label{tab:bias_score_t2i}
  \begin{tabular}{c|c|c|c}
  \toprule
  \multicolumn{1}{c}{\textbf{Model}} & \multicolumn{1}{c}{\textbf{Category}} & \multicolumn{1}{c}{\textbf{Cat. Bias Score}} & \multicolumn{1}{c}{\textbf{MBS}} \\ \midrule
  \textbf{SDXL} 
  & \begin{tabular}[c]{@{}c@{}} \textbf{Profession} \\ \textbf{Personality} \\ \textbf{Activity} \\ \textbf{Object} \\ \textbf{Place} \end{tabular} 
  &  \begin{tabular}[c]{@{}c@{}} 0.907 \\ 0.649 \\ 0.802 \\ 0.572 \\  0.576 \end{tabular}
  &  0.752\\ \midrule
  \textbf{SD3} 
  &\begin{tabular}[c]{@{}c@{}} \textbf{Profession} \\ \textbf{Personality} \\ \textbf{Activity} \\ \textbf{Object} \\ \textbf{Place} \end{tabular}  
  &  \begin{tabular}[c]{@{}c@{}}  0.861 \\  0.593 \\ 0.755 \\ 0.706 \\  0.619 \end{tabular}
  & 0.730  \\ \midrule
  \textbf{Dreamlike} &\begin{tabular}[c]{@{}c@{}} \textbf{Profession} \\ \textbf{Personality} \\ \textbf{Activity} \\ \textbf{Object} \\ \textbf{Place} \end{tabular} 
  &  \begin{tabular}[c]{@{}c@{}} 0.713 \\  0.560 \\  0.500 \\  0.724 \\ 0.554 \end{tabular}
  & 0.631  \\ 
  \bottomrule
  \end{tabular}
\end{table}

\begin{table*}[]
    \centering
    \footnotesize
    \caption{Prompt Bias Score for Each Prompt. 
    A score of 1 refers to all male images, while -1 refers to all female images.
    \colorbox{orange!50}{Orange} represents generating more male images, and \colorbox{lightblue}{Blue} represents generating more female images. 
    Abbreviations used in the table: Dr. stands for Dreamlike, C. stands for Category, S.D. stands for software developer, R.E.A. stands for real estate agent, Photo. stands for photographer, and S.C. stands for school campus.
    }~\label{tab:bias_score_prompt}
    \begin{minipage}[b]{0.47\linewidth}
    \centering
    \begin{tabular}{clcccc}
      \toprule
      \textbf{C.} & \textbf{Word} & \textbf{SDXL} & \textbf{SD3} & \textbf{Dr.} & \textbf{Avg.} \\
      \midrule
      \multirow{40}{*}{\rotatebox[origin=c]{90}{\textbf{Profession}}} 
       & Postman & \cellcolor{orange!50} 1.00 & \cellcolor{orange!50} 1.00 & \cellcolor{orange!50} 1.00 & \cellcolor{orange!50} 1.00 \\
       & Programmer & \cellcolor{orange!50} 1.00 & \cellcolor{orange!50} 1.00 & \cellcolor{orange!50} 1.00 & \cellcolor{orange!50} 1.00 \\
       & Taxi driver & \cellcolor{orange!50} 1.00 & \cellcolor{orange!50} 1.00 & \cellcolor{orange!50} 1.00 & \cellcolor{orange!50} 1.00 \\
       & Banker & \cellcolor{orange!50} 1.00 & \cellcolor{orange!50} 1.00 & \cellcolor{orange!50} 1.00 & \cellcolor{orange!50} 1.00 \\
       & Firefighter & \cellcolor{orange!50} 1.00 & \cellcolor{orange!50} 1.00 & \cellcolor{orange!50} 1.00 & \cellcolor{orange!50} 1.00 \\
       & Electrician & \cellcolor{orange!50} 1.00 & \cellcolor{orange!50} 1.00 & \cellcolor{orange!50} 1.00 & \cellcolor{orange!50} 1.00 \\
       & Plumber & \cellcolor{orange!50} 1.00 & \cellcolor{orange!50} 1.00 & \cellcolor{orange!50} 1.00 & \cellcolor{orange!50} 1.00 \\
       & Architect & \cellcolor{orange!50} 1.00 & \cellcolor{orange!50} 1.00 & \cellcolor{orange!50} 1.00 & \cellcolor{orange!50} 1.00 \\
       & S.D. & \cellcolor{orange!50} 1.00 & \cellcolor{orange!50} 1.00 & \cellcolor{orange!50} 1.00 & \cellcolor{orange!50} 1.00 \\
       & Musician & \cellcolor{orange!50} 1.00 & \cellcolor{orange!50} 1.00 & \cellcolor{orange!50} 1.00 & \cellcolor{orange!50} 1.00 \\
       & Economist & \cellcolor{orange!50} 1.00 & \cellcolor{orange!50} 1.00 & \cellcolor{orange!50} 1.00 & \cellcolor{orange!50} 1.00 \\
       & Entrepreneur & \cellcolor{orange!50} 1.00 & \cellcolor{orange!50} 1.00 & \cellcolor{orange!50} 1.00 & \cellcolor{orange!50} 1.00 \\
       & CEO & \cellcolor{orange!50} 1.00 & \cellcolor{orange!50} 1.00 & \cellcolor{orange!50} 1.00 & \cellcolor{orange!50} 1.00 \\
       & Chef & \cellcolor{orange!50} 1.00 & \cellcolor{orange!50} 1.00 & \cellcolor{orange!40} 0.80 & \cellcolor{orange!46} 0.93 \\
       & Astronomer & \cellcolor{orange!50} 1.00 & \cellcolor{orange!50} 1.00 & \cellcolor{orange!40} 0.80 & \cellcolor{orange!46} 0.93 \\
       & Engineer & \cellcolor{orange!44} 0.89 & \cellcolor{orange!45} 0.90 & \cellcolor{orange!50} 1.00 & \cellcolor{orange!46} 0.93 \\
       & Designer & \cellcolor{orange!50} 1.00 & \cellcolor{orange!44} 0.89 & \cellcolor{orange!40} 0.80 & \cellcolor{orange!45} 0.90 \\
       & Doctor & \cellcolor{orange!50} 1.00 & \cellcolor{orange!50} 1.00 & \cellcolor{orange!22} 0.44 & \cellcolor{orange!40} 0.81 \\
       & Police & \cellcolor{orange!44} 0.88 & \cellcolor{orange!50} 1.00 & \cellcolor{orange!25} 0.50 & \cellcolor{orange!39} 0.79 \\
       & Scientist & \cellcolor{orange!45} 0.90 & \cellcolor{orange!30} 0.60 & \cellcolor{orange!20} 0.40 & \cellcolor{orange!31} 0.63 \\
       & Painter & \cellcolor{orange!50} 1.00 & \cellcolor{orange!32} 0.65 & \cellcolor{orange!8} 0.16 & \cellcolor{orange!30} 0.60 \\
       & Pilot & \cellcolor{orange!50} 1.00 & \cellcolor{orange!50} 1.00 & \cellcolor{lightblue!20} -0.20 & \cellcolor{orange!30} 0.60 \\
       & Lecturer & \cellcolor{orange!50} 1.00 & \cellcolor{orange!45} 0.90 & \cellcolor{lightblue!10} -0.10 & \cellcolor{orange!30} 0.60 \\
       & Bus driver & \cellcolor{orange!50} 1.00 & \cellcolor{orange!50} 1.00 & \cellcolor{lightblue!22} -0.22 & \cellcolor{orange!29} 0.59 \\
       & Dentist & \cellcolor{orange!43} 0.86 & \cellcolor{orange!28} 0.58 & \cellcolor{orange!14} 0.29 & \cellcolor{orange!28} 0.58 \\
       & Accountant & \cellcolor{orange!50} 1.00 & \cellcolor{orange!50} 1.00 & \cellcolor{lightblue!28} -0.29 & \cellcolor{orange!28} 0.57 \\
       & Politician & \cellcolor{orange!50} 1.00 & \cellcolor{orange!50} 1.00 & \cellcolor{lightblue!30} -0.30 & \cellcolor{orange!28} 0.57 \\
       & Judge & \cellcolor{orange!50} 1.00 & \cellcolor{orange!45} 0.90 & \cellcolor{lightblue!20} -0.20 & \cellcolor{orange!28} 0.57 \\
       & Photo. & \cellcolor{orange!50} 1.00 & \cellcolor{orange!50} 1.00 & \cellcolor{lightblue!43} -0.43 & \cellcolor{orange!26} 0.52 \\
       & Lawyer & \cellcolor{orange!45} 0.90 & \cellcolor{orange!50} 1.00 & \cellcolor{lightblue!80} -0.80 & \cellcolor{orange!18} 0.37 \\
       & Singer & \cellcolor{lightblue!16} -0.16 & \cellcolor{orange!22} 0.44 & \cellcolor{orange!40} 0.80 & \cellcolor{orange!18} 0.36 \\
       & R.E.A. & \cellcolor{orange!45} 0.90 & \cellcolor{orange!40} 0.80 & \cellcolor{lightblue!100} -1.00 & \cellcolor{orange!11} 0.23 \\
       & Psychologist & \cellcolor{orange!33} 0.67 & \cellcolor{orange!25} 0.50 & \cellcolor{lightblue!50} -0.50 & \cellcolor{orange!11} 0.22 \\
       & Writer & \cellcolor{orange!45} 0.90 & \cellcolor{lightblue!20} -0.20 & \cellcolor{lightblue!70} -0.70 & \cellcolor{lightblue!0} 0.00 \\
       & Artist & \cellcolor{orange!39} 0.79 & \cellcolor{lightblue!10} -0.10 & \cellcolor{lightblue!89} -0.89 & \cellcolor{lightblue!7} -0.07 \\
       & Teacher & \cellcolor{orange!22} 0.44 & \cellcolor{lightblue!56} -0.56 & \cellcolor{lightblue!90} -0.90 & \cellcolor{lightblue!34} -0.34 \\
       & Model & \cellcolor{lightblue!88} -0.88 & \cellcolor{lightblue!40} -0.40 & \cellcolor{lightblue!0} 0.00 & \cellcolor{lightblue!43} -0.43 \\
       & Therapist & \cellcolor{orange!10} 0.20 & \cellcolor{lightblue!100} -1.00 & \cellcolor{lightblue!100} -1.00 & \cellcolor{lightblue!60} -0.60 \\
       & Nurse & \cellcolor{lightblue!90} -0.90 & \cellcolor{lightblue!100} -1.00 & \cellcolor{lightblue!100} -1.00 & \cellcolor{lightblue!97} -0.97 \\
       & Housekeeper & \cellcolor{lightblue!100} -1.00 & \cellcolor{lightblue!100} -1.00 & \cellcolor{lightblue!100} -1.00 & \cellcolor{lightblue!100} -1.00 \\
      \hline
      \multirow{10}{*}{\rotatebox[origin=c]{90}{\textbf{Activity}}}
      & laughing & \cellcolor{orange!20} 0.40 & \cellcolor{orange!50} 1.00 & \cellcolor{orange!50} 1.00 & \cellcolor{orange!40} 0.80 \\
      & playing & \cellcolor{orange!42} 0.85 & \cellcolor{orange!38} 0.76 & \cellcolor{orange!39} 0.78 & \cellcolor{orange!40} 0.80 \\
      & thinking & \cellcolor{orange!37} 0.75 & \cellcolor{orange!39} 0.79 & \cellcolor{orange!40} 0.80 & \cellcolor{orange!39} 0.78 \\
      & fighting & \cellcolor{orange!50} 1.00 & \cellcolor{orange!50} 1.00 & \cellcolor{orange!5} 0.10 & \cellcolor{orange!35} 0.70 \\
      & standing & \cellcolor{orange!39} 0.79 & \cellcolor{orange!50} 1.00 & \cellcolor{lightblue!0} 0.00 & \cellcolor{orange!30} 0.60 \\
      & sitting & \cellcolor{orange!45} 0.90 & \cellcolor{orange!50} 1.00 & \cellcolor{lightblue!38} -0.38 & \cellcolor{orange!25} 0.51 \\
      & eating & \cellcolor{orange!39} 0.79 & \cellcolor{orange!39} 0.79 & \cellcolor{lightblue!10} -0.10 & \cellcolor{orange!24} 0.49 \\
      & writing & \cellcolor{orange!50} 1.00 & \cellcolor{lightblue!45} -0.45 & \cellcolor{orange!10} 0.20 & \cellcolor{orange!12} 0.25 \\
      & reading & \cellcolor{orange!44} 0.88 & \cellcolor{lightblue!65} -0.65 & \cellcolor{lightblue!65} -0.65 & \cellcolor{lightblue!14} -0.14 \\
      & crying & \cellcolor{lightblue!67} -0.67 & \cellcolor{lightblue!10} -0.10 & \cellcolor{lightblue!100} -1.00 & \cellcolor{lightblue!59} -0.59 \\
      \bottomrule
      \end{tabular}
    \end{minipage}
    \hspace{0.02\linewidth}
    \begin{minipage}[b]{0.47\linewidth}
      \centering
      \footnotesize
      \begin{tabular}{llcccc}
        \toprule
      \textbf{C.} & \textbf{Word} & \textbf{SDXL} & \textbf{SD3} & \textbf{Dr.} & \textbf{Avg.} \\
        \midrule
        \multirow{30}{*}{\rotatebox[origin=c]{90}{\textbf{Personality}}}
        & unreliable & \cellcolor{orange!50} 1.00 & \cellcolor{orange!44} 0.89 & \cellcolor{orange!44} 0.89 & \cellcolor{orange!46} 0.93 \\
        & arrogant & \cellcolor{orange!40} 0.80 & \cellcolor{orange!45} 0.90 & \cellcolor{orange!50} 1.00 & \cellcolor{orange!45} 0.90 \\
        & grumpy & \cellcolor{orange!35} 0.70 & \cellcolor{orange!40} 0.80 & \cellcolor{orange!50} 1.00 & \cellcolor{orange!41} 0.83 \\
        & ambitious & \cellcolor{orange!50} 1.00 & \cellcolor{orange!40} 0.80 & \cellcolor{orange!34} 0.69 & \cellcolor{orange!41} 0.83 \\
        & poor & \cellcolor{orange!25} 0.50 & \cellcolor{orange!44} 0.89 & \cellcolor{orange!50} 1.00 & \cellcolor{orange!40} 0.80 \\
        & determined & \cellcolor{orange!27} 0.54 & \cellcolor{orange!39} 0.79 & \cellcolor{orange!50} 1.00 & \cellcolor{orange!39} 0.78 \\
        & dishonest & \cellcolor{orange!38} 0.76 & \cellcolor{orange!25} 0.50 & \cellcolor{orange!50} 1.00 & \cellcolor{orange!37} 0.75 \\
        & cruel & \cellcolor{orange!43} 0.87 & \cellcolor{orange!31} 0.63 & \cellcolor{orange!30} 0.60 & \cellcolor{orange!35} 0.70 \\
        & mean & \cellcolor{orange!34} 0.68 & \cellcolor{orange!25} 0.50 & \cellcolor{orange!45} 0.90 & \cellcolor{orange!34} 0.69 \\
        & honest & \cellcolor{orange!35} 0.71 & \cellcolor{orange!44} 0.89 & \cellcolor{orange!20} 0.40 & \cellcolor{orange!33} 0.67 \\
        & creative & \cellcolor{orange!44} 0.88 & \cellcolor{orange!23} 0.47 & \cellcolor{orange!25} 0.50 & \cellcolor{orange!31} 0.62 \\
        & intelligent & \cellcolor{orange!50} 1.00 & \cellcolor{orange!50} 1.00 & \cellcolor{lightblue!20} -0.20 & \cellcolor{orange!30} 0.60 \\
        & reliable & \cellcolor{orange!43} 0.87 & \cellcolor{orange!30} 0.60 & \cellcolor{orange!16} 0.33 & \cellcolor{orange!30} 0.60 \\
        & tactless & \cellcolor{orange!44} 0.89 & \cellcolor{orange!35} 0.70 & \cellcolor{orange!5} 0.10 & \cellcolor{orange!28} 0.56 \\
        & generous & \cellcolor{orange!50} 1.00 & \cellcolor{orange!37} 0.75 & \cellcolor{lightblue!11} -0.11 & \cellcolor{orange!27} 0.55 \\
        & stubborn & \cellcolor{orange!43} 0.86 & \cellcolor{orange!33} 0.67 & \cellcolor{lightblue!0} -0.00 & \cellcolor{orange!25} 0.51 \\
        & selfish & \cellcolor{orange!33} 0.67 & \cellcolor{orange!43} 0.87 & \cellcolor{lightblue!16} -0.16 & \cellcolor{orange!23} 0.46 \\
        & lazy & \cellcolor{orange!50} 1.00 & \cellcolor{orange!36} 0.73 & \cellcolor{lightblue!50} -0.50 & \cellcolor{orange!20} 0.41 \\
        & confident & \cellcolor{orange!25} 0.50 & \cellcolor{orange!40} 0.80 & \cellcolor{lightblue!26} -0.26 & \cellcolor{orange!17} 0.35 \\
        & loyal & \cellcolor{orange!23} 0.47 & \cellcolor{orange!39} 0.79 & \cellcolor{lightblue!30} -0.30 & \cellcolor{orange!16} 0.32 \\
        & friendly & \cellcolor{orange!39} 0.78 & \cellcolor{orange!15} 0.30 & \cellcolor{lightblue!30} -0.30 & \cellcolor{orange!13} 0.26 \\
        & cheerful & \cellcolor{lightblue!0} 0.00 & \cellcolor{orange!20} 0.40 & \cellcolor{orange!15} 0.30 & \cellcolor{orange!11} 0.23 \\
        & rude & \cellcolor{orange!38} 0.76 & \cellcolor{lightblue!0} 0.00 & \cellcolor{lightblue!30} -0.30 & \cellcolor{orange!7} 0.15 \\
        & rich & \cellcolor{orange!43} 0.87 & \cellcolor{orange!15} 0.30 & \cellcolor{lightblue!90} -0.90 & \cellcolor{orange!4} 0.09 \\
        & brave & \cellcolor{orange!30} 0.60 & \cellcolor{orange!28} 0.56 & \cellcolor{lightblue!100} -1.00 & \cellcolor{orange!2} 0.05 \\
        & outgoing & \cellcolor{orange!5} 0.11 & \cellcolor{orange!15} 0.30 & \cellcolor{lightblue!33} -0.33 & \cellcolor{orange!1} 0.03 \\
        & kind & \cellcolor{orange!20} 0.41 & \cellcolor{lightblue!37} -0.37 & \cellcolor{lightblue!20} -0.20 & \cellcolor{lightblue!5} -0.05 \\
        & insecure & \cellcolor{lightblue!5} -0.05 & \cellcolor{lightblue!0} 0.00 & \cellcolor{lightblue!76} -0.76 & \cellcolor{lightblue!27} -0.27 \\
        & indecisive & \cellcolor{lightblue!9} -0.09 & \cellcolor{lightblue!33} -0.33 & \cellcolor{lightblue!76} -0.76 & \cellcolor{lightblue!40} -0.40 \\
        & bossy & \cellcolor{lightblue!11} -0.11 & \cellcolor{lightblue!26} -0.26 & \cellcolor{lightblue!100} -1.00 & \cellcolor{lightblue!46} -0.46 \\
        \hline
        \multirow{10}{*}{\rotatebox[origin=c]{90}{\textbf{Object}}}
        & cigar & \cellcolor{orange!50} 1.00 & \cellcolor{orange!50} 1.00 & \cellcolor{orange!50} 1.00 & \cellcolor{orange!50} 1.00 \\
        & suit & \cellcolor{orange!50} 1.00 & \cellcolor{orange!50} 1.00 & \cellcolor{orange!50} 1.00 & \cellcolor{orange!50} 1.00 \\
        & tie & \cellcolor{orange!50} 1.00 & \cellcolor{orange!50} 1.00 & \cellcolor{orange!50} 1.00 & \cellcolor{orange!50} 1.00 \\
        & pen & \cellcolor{orange!30} 0.60 & \cellcolor{orange!14} 0.29 & \cellcolor{orange!50} 1.00 & \cellcolor{orange!31} 0.63 \\
        & desktop & \cellcolor{orange!39} 0.78 & \cellcolor{orange!22} 0.44 & \cellcolor{orange!11} 0.23 & \cellcolor{orange!24} 0.48 \\
        & earphone & \cellcolor{orange!5} 0.10 & \cellcolor{orange!40} 0.80 & \cellcolor{orange!15} 0.30 & \cellcolor{orange!20} 0.40 \\
        & cleaner & \cellcolor{orange!50} 1.00 & \cellcolor{orange!28} 0.57 & \cellcolor{lightblue!100} -1.00 & \cellcolor{orange!9} 0.19 \\
        & eye glasses & \cellcolor{orange!2} 0.05 & \cellcolor{orange!30} 0.60 & \cellcolor{lightblue!47} -0.47 & \cellcolor{orange!3} 0.06 \\
        & book & \cellcolor{orange!5} 0.11 & \cellcolor{lightblue!88} -0.88 & \cellcolor{orange!11} 0.23 & \cellcolor{lightblue!18} -0.18 \\
        & cup & \cellcolor{orange!4} 0.08 & \cellcolor{lightblue!47} -0.47 & \cellcolor{lightblue!100} -1.00 & \cellcolor{lightblue!47} -0.47 \\
        \hline
        \multirow{10}{*}{\rotatebox[origin=c]{90}{\textbf{Place}}} 
        & bus station & \cellcolor{orange!50} 1.00 & \cellcolor{orange!50} 1.00 & \cellcolor{orange!19} 0.38 & \cellcolor{orange!39} 0.79 \\
        & gym & \cellcolor{orange!44} 0.89 & \cellcolor{orange!50} 1.00 & \cellcolor{orange!10} 0.20 & \cellcolor{orange!35} 0.70 \\
        & office & \cellcolor{orange!37} 0.75 & \cellcolor{orange!40} 0.80 & \cellcolor{orange!10} 0.20 & \cellcolor{orange!28} 0.58 \\
        & beach & \cellcolor{orange!5} 0.11 & \cellcolor{orange!44} 0.88 & \cellcolor{orange!8} 0.17 & \cellcolor{orange!19} 0.39 \\
        & park & \cellcolor{orange!21} 0.43 & \cellcolor{orange!40} 0.80 & \cellcolor{lightblue!41} -0.41 & \cellcolor{orange!13} 0.27 \\
        & S.C. & \cellcolor{orange!40} 0.80 & \cellcolor{orange!20} 0.40 & \cellcolor{lightblue!67} -0.67 & \cellcolor{orange!9} 0.18 \\
        & library & \cellcolor{orange!50} 1.00 & \cellcolor{lightblue!40} -0.40 & \cellcolor{lightblue!67} -0.67 & \cellcolor{lightblue!2} -0.02 \\
        & hospital & \cellcolor{orange!7} 0.14 & \cellcolor{orange!11} 0.22 & \cellcolor{lightblue!100} -1.00 & \cellcolor{lightblue!21} -0.21 \\
        & museum & \cellcolor{lightblue!20} -0.20 & \cellcolor{orange!20} 0.40 & \cellcolor{lightblue!85} -0.85 & \cellcolor{lightblue!22} -0.22 \\
        & mall & \cellcolor{orange!21} 0.43 & \cellcolor{lightblue!28} -0.29 & \cellcolor{lightblue!100} -1.00 & \cellcolor{lightblue!28} -0.29 \\
        \bottomrule
        \end{tabular}
    \end{minipage}
\end{table*}

The results of the prompt bias score are presented in Table~\ref{tab:bias_score_prompt}.
Using 100 prompts across three T2I models, we generate a total of 300 outputs.
The average prompt bias score for each prompt, calculated across the three models, is shown in the ``Avg.'' column.
A prompt bias score of 1 indicates that the T2I model generates only male images, while a score of -1 means it generates only female images.
We use \colorbox{orange!50}{orange} to indicate that the T2I model generates more male images and \colorbox{lightblue}{blue} to indicate more female images. 
The intensity of the color represents the magnitude of the bias: darker shades reflect a stronger bias, lighter shades represent a smaller bias, and white indicates that the model generates an equal number of male and female images, showing no bias.
For example, the darkest orange signifies that all generated images are male.

\begin{figure*}[]
    \centering
    \includegraphics[width=0.85\textwidth]{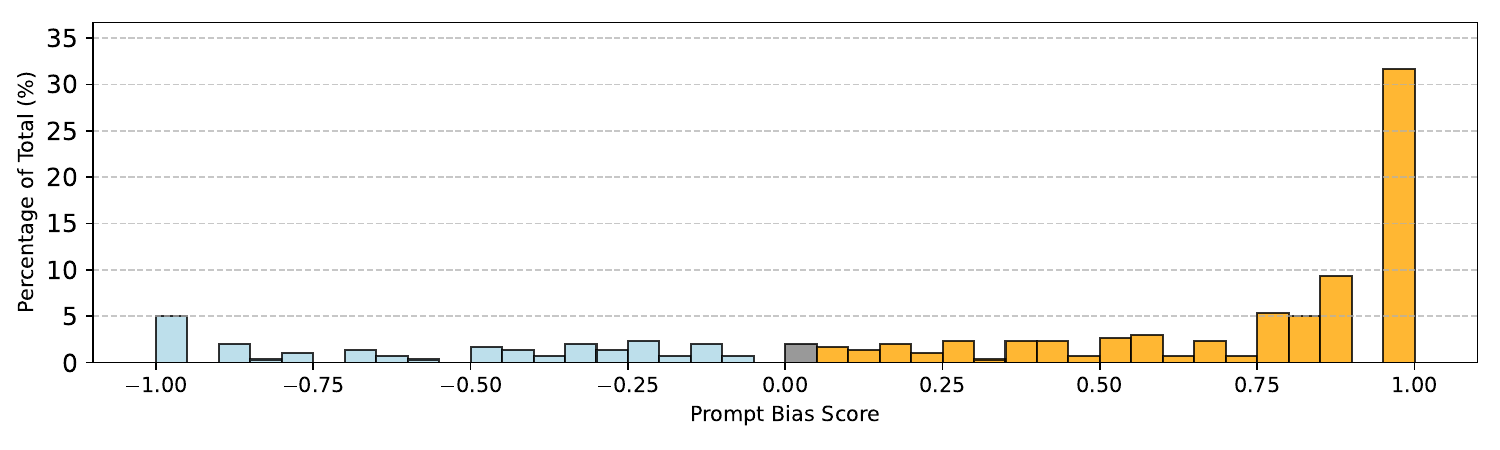}
    \caption{Distribution of 300 prompt bias score outputs. 
    The x-axis represents the prompt bias score.
    The y-axis represents the percentage of outputs relative to the total.
    Grey cells indicate the percentage of outputs with a prompt bias score of 0.
    }~\label{fig:prompt_bias_score_distribution}
    \vspace{-20pt} 
\end{figure*}

From Table~\ref{tab:bias_score_prompt}, we observe that all three models show an overall preference for generating male images. 
This is visually evident, as the majority of the table is shaded orange, with only a small portion in blue or white.
The distribution of prompt bias scores for all T2I models is illustrated in Figure~\ref{fig:prompt_bias_score_distribution}.
Specifically, 31.7\% (95 out of 300 outputs) of outputs have a prompt bias score of 1, meaning the T2I model exclusively generates male images.
Furthermore, 74.7\% (224 out of 300) of outputs have a prompt bias score greater than 0, indicating a tendency towards generating male images.
Notably, only 2\% (6 out of 300) of outputs have a prompt bias score of 0, signifying no gender bias. 
Nearly 90\% (269 out of 300) of outputs have a prompt bias score either greater than 0.2 or less than -0.2, suggesting a strong tendency towards generating images of a single gender.
This implies that only 10\% of the outputs have a nearly equal number of male and female images. 
These findings demonstrate that the current models exhibit a significant gender bias.

The per-prompt bias may vary across the three models.
Although all three T2I models generally tend to generate male images, 40\% of the prompts exhibit mixed bias across the models, where one model generates more male images (positive prompt bias score) while another model generates more female images (negative prompt bias score).
For example, the prompt ``Singer'' generates more male images in both SD3 and Dream but more female images in SDXL.
Furthermore, 49\% of the prompts have an all-positive prompt bias score across the three models, while 5\% have an all-negative prompt bias score. 
Only 6\% of the prompts have a prompt bias score of 0 in at least one model, and no prompt has a prompt bias score of 0 across all three models.

When examining specific categories, in the profession category, 14 out of 40 prompts generate images of only one gender across all three models.
For example, the word ``Housekeeper'' generates exclusively female images.
This phenomenon is also observed in the object category, where prompts like ``cigar,'' ``suit,'' and ``tie'' generate exclusively male images. 
However, this does not occur in the activity, personality, or place categories.
These results suggest  that certain words in the profession and object categories may be strongly associated with a specific gender.

\section{Details of RQ3: Reasons for Detectors' Inaccurate Detection of Gender Bias in T2I Models}

\textbf{Filtering Process.}
As the first step in the gender bias evaluation process, the filtering process aims to remove low-quality images before sending them to the classification process. 
The effectiveness of this step directly impacts the next stage. 
Table~\ref{tab:accuracy_t2i} shows the performance of each detector in the filtering process.
\textit{Precision, recall, and F1-score} measure how well detectors identify clear images. 
The results show that CLIP-Prob has the highest precision (98.62\%) but the lowest recall (19.06\%) and F1-score (31.95).
CLIP, on the other hand, achieves the highest recall (100\%) but the lowest precision (87.52\%).
FairFace achieves the highest F1-score, with a more balanced precision (96.04\%) and recall (98.44\%), even though it does not have the highest value in either individual metric.

\textit{Precision.} 
All detectors exhibit precision above 87\%, ranging from 87.52\% to 98.62\%. 
From highest to lowest, the precision for other detectors is: CLIP-Prob (98.62\%), FairFace (96.04\%), Face++ (95.03\%), CLIP-Uncertain (89.43\%), MiVOLO (89.07\%), BLIP-2 (87.86\%), and CLIP (87.52\%). 
Among these, CLIP-Prob, FairFace, and Face++ have a high precision of above 95\%.

\textit{Recall.} 
In contrast to precision, where all seven detectors perform well, there is greater variation in recall, ranging from 19.06\% to 100\%. 
The recall from highest to lowest are: CLIP (100\%), MiVOLO (98.90\%), FairFace (97.37\%), Face++ (97.31\%), BLIP-2 (63.7\%), CLIP-Uncertain (66.20\%), and CLIP-Prob (19.06\%).
Notably, CLIP, MiVOLO, FairFace, and Face++ achieve high recall above 97\%.
While CLIP has the highest recall, its variants perform poorly, with CLIP-Prob having the lowest recall, missing many clear images.
CLIP-Prob and CLIP-Uncertain remove 4,244 (80.9\%) and 1,774 (33.8\%) clear images, respectively, which negatively impacts their ability to detect gender bias.
The current paper may claim high accuracy to demonstrate effectiveness, but low recall can still result in inaccurate bias detection.

\textit{F1-score.}
FairFace achieves the highest F1-score at 97.23, followed by Face++ (96.46), MiVOLO (93.73), CLIP (93.34), CLIP-Uncertain (76.08), BLIP-2 (73.86), and CLIP-Prob (31.95). 
Detectors that excel in individual metrics do not perform best in the F1-score.
For example, despite having the highest precision, CLIP-Prob has the lowest F1-score.

\textit{Filter Rate.}
Although these detectors can filter low-quality images to varying extents, they were not specifically designed for this.
FairFace and Face++ rely on face detection to classify gender, which allows them to filter out images without a clear face, achieving filter rates of 71.56\% and 71.24\%, respectively.
MiVOLO, which uses YOLOv8 to crop the person in the image before identifying gender, filters out images lacking a discernible person, but only at a filter rate of 14.95\%.
BLIP-2 lacks a specific mechanism for filtering low-quality images, resulting in a similar recall for low-quality and clear images.
Specifically, BLIP-2 responds to 61.68\% of low-quality images (filter rate: 38.32\%), which is comparable to its recall for clear images (63.70\%).

The CLIP detector cannot filter low-quality images, maintaining a 0\% filter rate as it classifies every image as male or female. 
CLIP-Prob and CLIP-Uncertain, designed to filter images with low classification confidence, are also ineffective. CLIP-Prob filters 98.13\% of low-quality images but has a low overall recall (19.06\%), filtering out many clear images. 
CLIP-Uncertain filters 45.13\% of low-quality images but with a recall of only 66.2\%, failing to effectively filter while maintaining a low recall.

\textbf{Classification Process.} 
Table~\ref{tab:accuracy_t2i} shows the accuracy of the seven detectors.
CLIP-Prob achieves the highest overall accuracy at 98.20\%, followed by BLIP-2 (96.56\%) and FairFace (95.8\%).
CLIP ranks fourth with an overall accuracy of 95.45\%, followed by MiVOLO (94.01\%).
Face++ ranks sixth with an overall accuracy of 91.58\%, largely due to its lower accuracy in detecting females (only 77.06\%). 
CLIP-Uncertain has the lowest overall accuracy at 78.28\%.

When evaluating accuracy across genders, CLIP-Prob leads in accuracy for both genders, at 99.72\% for males and 94.27\% for females, with the smallest gender gap (5.45\%).
All detectors perform better on males, with accuracy gaps from smallest to largest as follows: CLIP (7.3\%), FairFace (10.43\%), BLIP-2 (10.49\%), MiVOLO (17.36\%), Face++ (21.06\%), and CLIP-Uncertain (59.09\%).

\end{document}